\documentclass[runningheads]{llncs}

\usepackage[T1]{fontenc}
\usepackage{graphicx}
\usepackage{amsmath,amssymb}
\usepackage{booktabs}
\usepackage{url}
\begin{document}

\title{CRIL-U-Net: Compact Ratio-Interaction Learning for Focal Cortical Dysplasia Segmentation from T1w and FLAIR MRI}
\titlerunning{CRIL-U-Net for FCD Segmentation}

\author{Soumen Ghosh\inst{1}\orcidID{0000-0002-8360-5309}\thanks{Corresponding author: soumen.ghosh@uq.edu.au} \and
Subhojit Mandal\inst{2} \and
Tilottama Goswami\inst{3} \and
Amit Soni Arya\inst{4} \and
John Phamnguyen\inst{1,5} \and
Rajat Vashistha\inst{1}}
\authorrunning{S. Ghosh et al.}

\institute{The University of Queensland, Brisbane, Australia \and
IIT Madras, Chennai, India \and
University College of Engineering, Osmania University, Hyderabad, India \and
Bennett University, Greater Noida, India \and
Royal Brisbane and Women's Hospital, Brisbane, Australia\\}
\maketitle
\vspace{-0.5cm}
\begin{abstract}
Focal cortical dysplasia (FCD) type II is an important cause of drug-resistant focal epilepsy, but its small size and subtle magnetic resonance imaging (MRI) appearance make automated segmentation difficult. Conventional multimodal networks concatenate T1-weighted (T1w) and fluid-attenuated inversion recovery (FLAIR) images and rely on subsequent layers to learn their relationships implicitly. We propose Compact Ratio-Interaction Learning U-Net (CRIL-U-Net), a 3D U-Net extension incorporating a Compact Ratio-Interaction Learning module that combines local spatial features, voxel-wise cross-modal mixing, and bidirectional ratio-inspired interactions. CRIL-U-Net was compared with a 3D U-Net and a U-Net with input-level self-attention using five-fold cross-validation on 85 FCD subjects and 25 healthy controls. Each architecture was trained independently with Dice-binary cross-entropy (Dice-BCE) and Focal Tversky-Focal (FTF) losses. With FTF, CRIL-U-Net achieved the highest mean Dice score ($0.196\pm0.262$), compared with $0.136\pm0.224$ for the U-Net and $0.135\pm0.214$ for the attention comparator. It localised 44 of 85 lesions, compared with 36 for the U-Net, and significantly outperformed both comparison architectures after false-discovery-rate correction. CRIL adds only 4,320 parameters to the backbone, indicating that compact cross-modal representation learning can improve FCD segmentation when combined with an imbalance-aware objective. The remaining 48.2\% miss rate nevertheless requires further methodological development and external validation.

\end{abstract}

\section{Introduction}

Epilepsy affects more than 65 million people worldwide, and focal cortical dysplasia (FCD) is a leading structural cause of drug-resistant focal epilepsy \cite{ngugi2010estimation,taylor1971focal}. Surgical resection can provide seizure freedom when the epileptogenic lesion is accurately localised and delineated \cite{ryvlin2014epilepsy,kim2011neuroimaging,harvey2015surgically}. However, FCD abnormalities are frequently small, heterogeneous, and only subtly distinguishable from surrounding tissue, making reliable segmentation from structural MRI difficult.

T1-weighted (T1w) MRI depicts anatomical features such as cortical thickening, abnormal gyration, and blurring of the grey-white matter junction, whereas fluid-attenuated inversion recovery (FLAIR) is sensitive to lesion-associated signal abnormalities \cite{bernasconi2001texture,gill2017automated,sisodiya2009focal}. These modalities are complementary, yet FCD can remain visually subtle or MRI-negative, motivating automated analysis \cite{spitzer2022interpretable,snell2026mri}. Deep networks, particularly U-Net and nnU-Net, have advanced medical image segmentation \cite{ronneberger2015u,isensee2021nnu}, but FCD remains challenging because lesions occupy a minute proportion of the brain, annotated datasets are small, and lesion appearance varies substantially across subjects.

Multimodal segmentation networks usually concatenate T1w and FLAIR channels. Although simple, concatenation places the burden of discovering useful modality relationships on subsequent convolutions. Attention mechanisms can emphasise informative features \cite{oktay2018attention}, but do not necessarily encode complementary intensity relationships between modalities. Fixed T1w/FLAIR and FLAIR/T1w representations have shown complementary value for FCD segmentation \cite{ghosh2026benchmarking}, suggesting that explicit cross-modal interactions may be beneficial.

We therefore propose CRIL-U-Net, which learns compact local, cross-modal, and bidirectional ratio-inspired representations before segmentation by a 3D U-Net, while adding only 4,320 trainable parameters (0.019\% relative to the baseline). To distinguish the value of this representation from increased complexity or attention alone, we compare 3D U-Net, CRIL-U-Net, and a lightweight input self-attention U-Net under identical preprocessing, data partitions, training settings, and two loss functions. This study evaluates how architectural design and imbalance-aware optimisation jointly influence FCD segmentation performance in this challenging small-lesion setting.

\section{Related Work}

Recent FCD studies have progressed from cortical classification to voxel-wise lesion segmentation, including multicentre deep-learning systems and nnU-Net-based approaches \cite{gill2021multicenter,thomas2020multi,ding2025automated,joshi2025nnu}. Several FCD segmentation studies have focused primarily on backbone design using predefined MRI inputs \cite{joshi2025nnu,thomas2020multi}, leaving representation design comparatively underexplored \cite{ghosh2026benchmarking}. Multimodal medical-image methods instead combine modalities through channel concatenation, feature fusion, or attention \cite{zhou2019review,wang2024multimodal}. These strategies can integrate modality-specific features but generally learn interactions implicitly.

Ratio-derived MRI maps offer a complementary representation by emphasising relative tissue contrast. A recent benchmark found that ratio maps performed poorly alone but added information when combined with T1w and FLAIR \cite{ghosh2026benchmarking}. Because fixed ratios cannot adapt to nonlinear or task-specific interactions, the proposed CRIL module learns ratio-inspired and cross-modal features jointly with the segmentation objective while retaining a conventional 3D U-Net backbone.

Attention U-Net gates spatially relevant encoder features \cite{oktay2018attention}, whereas transformer-based architectures such as TransUNet, Swin-UNet, UNETR, Swin UNETR, and UNETR++ model longer-range dependencies through self-attention \cite{chen2024transunet,cao2022swin,hatamizadeh2022unetr,hatamizadeh2021swin,shaker2024unetr++}. These approaches primarily alter the segmentation backbone. Our controlled design instead places either a compact self-attention comparator or CRIL before the same U-Net, allowing global contextual refinement to be distinguished from explicit cross-modal representation learning.

\section{Methods}

\subsection{Dataset and Preprocessing}

We used the public dataset of Schuch et al. \cite{schuch2023open}, containing 85 subjects with FCD and 85 neurologically normal controls. The FCD cohort included 35 females and 50 males (age $28.9\pm12.4$ years); controls had a mean age of $33.3\pm11.9$ years. All 85 FCD subjects and a random subset of 25 controls were included. Expert lesion masks were available for FCD subjects, while controls had empty masks.

T1w and FLAIR volumes were aligned using affine registration in ANTs. A T1w-derived brain mask was applied to both modalities, followed by independent N4 bias correction, resampling to $1\,\mathrm{mm}^3$ isotropic resolution, and modality-wise intensity normalisation. Ground-truth lesion volume was calculated from the resampled masks and expressed in mL.

\subsection{Problem Formulation}

For co-registered volumes $X_T,X_F\in\mathbb{R}^{H\times W\times D}$, the paired input is $X=[X_T,X_F]\in\mathbb{R}^{2\times H\times W\times D}$. The network produces a logit map $L$ and lesion probabilities $\hat{Y}$:
\begin{equation}
L=g_\theta(X),\qquad \hat{Y}=\operatorname{sigmoid}(L),\qquad
Y\in\{0,1\}^{H\times W\times D}
\end{equation}
where \(H\), \(W\), and \(D\) denote the three spatial dimensions.
The baseline uses $L=g_\theta(X)$; CRIL-U-Net uses $L=g_\theta(\phi_{\mathrm{CRIL}}(X))$, where $\phi_{\mathrm{CRIL}}$ produces a four-channel latent representation. The attention comparator instead refines $X$ while retaining two channels.

\subsection{Architectures}

\begin{figure}[t]
\centering
\includegraphics[width=0.94\textwidth]{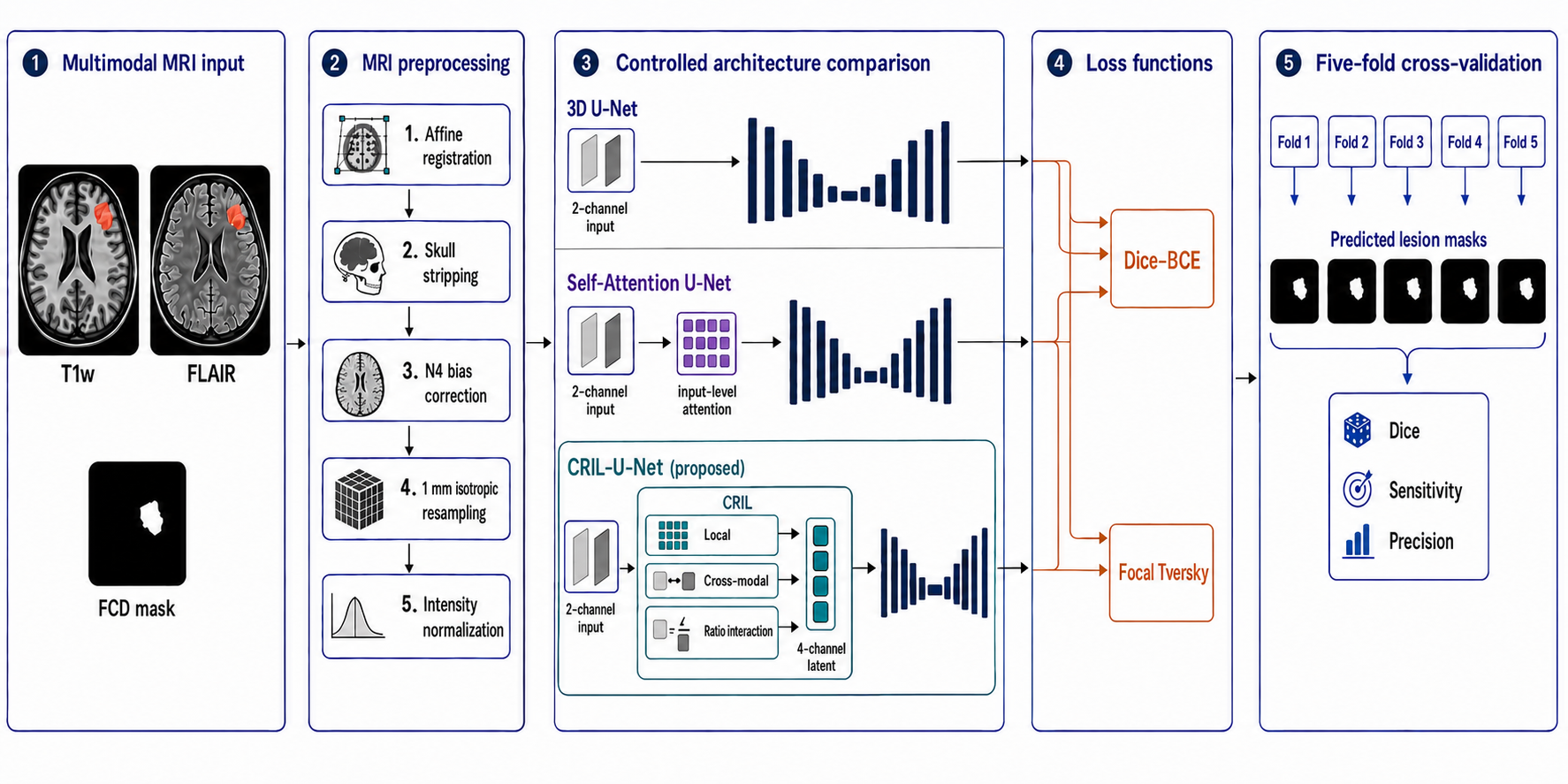}
\caption{Study framework and evaluated architectures. CRIL transforms paired T1w and FLAIR images into a compact cross-modal representation before the common 3D U-Net backbone.}
\label{fig:framework}
\end{figure}

Figure~\ref{fig:framework} summarises the three models. All used the same five-level 3D U-Net backbone, with encoder widths of 32, 64, 128, and 256 channels and 512 channels at the bottleneck. Each stage contained two $3^3$ convolutions followed by instance normalisation and LeakyReLU. Max pooling performed downsampling; transposed convolutions, skip concatenation, and paired convolution blocks formed the decoder. A final $1^3$ convolution produced lesion logits.

\paragraph{3D U-Net.}
The baseline directly processed the concatenated T1w and FLAIR channels.

\paragraph{Self-Attention U-Net.}
A lightweight self-attention block refined the two-channel input before the same backbone. A $1^3$ convolution produced 16-dimensional embeddings, adaptive pooling generated a $6^3$ grid (216 tokens), and four-head self-attention and a feed-forward layer were applied with pre-normalisation, residual connections, and dropout of 0.1. The representation was trilinearly upsampled, projected to two channels, and added to the original input.

\paragraph{CRIL-U-Net.}
For co-registered input $X=[X_T,X_F]$, CRIL constructs local, cross-modal, and ratio-interaction streams. The numerically stable bidirectional ratio representation is defined as:
\begin{equation}
X_{\mathrm{ratio}}=\left[
\operatorname{clip}\!\left(\frac{X_T}{s(X_F)},-r_{\max},r_{\max}\right),
\operatorname{clip}\!\left(\frac{X_F}{s(X_T)},-r_{\max},r_{\max}\right)
\right],
\end{equation}
where $s(A)=\epsilon$ when $|A|<\epsilon$ and $s(A)=A$ otherwise, with $\epsilon=10^{-3}$ and $r_{\max}=10$. The three streams are
\begin{align}
F_{\mathrm{local}}&=\sigma(\mathrm{IN}(W_{3^3}^{l}*X)),&
F_{\mathrm{cross}}&=\sigma(\mathrm{IN}(W_{1^3}^{c}*X)),\nonumber\\
F_{\mathrm{ratio}}&=\sigma(\mathrm{IN}(W_{3^3}^{r}*X_{\mathrm{ratio}})),&
Z&=h_{\mathrm{fuse}}([F_{\mathrm{local}},F_{\mathrm{cross}},F_{\mathrm{ratio}}])
\end{align}
where IN denotes instance normalisation, \(\sigma\) denotes LeakyReLU, and \(*\) denotes convolution.
Intuitively, the local stream captures short-range anatomical patterns, the cross-modal stream learns interactions between T1w and FLAIR intensities, and the ratio stream emphasises relative contrast differences that may make subtle lesions more distinguishable. Their fused representation is then supplied to the common U-Net backbone for segmentation.
Each stream produces 16 channels. Two $1^3$ fusion convolutions compress the 48 concatenated channels first to 16 and then to a four-channel latent representation $Z$, with instance normalisation and LeakyReLU after each convolution. The module and U-Net are trained end-to-end. The baseline, attention, and CRIL models contain 22,573,249, 22,575,537, and 22,577,569 trainable parameters, respectively; CRIL therefore adds only 4,320 parameters (0.019\%).

\subsection{Training and Loss Functions}

Five-fold stratified cross-validation preserved the FCD/control distribution. In each iteration, one fold was held out for testing; 15\% of the remaining subjects formed an internal validation subset for checkpoint selection. Fixed seed 42 and identical partitions were used across all configurations. Models were trained for 500 epochs with $96^3$ patches, batch size four, Adam optimisation, learning rate $10^{-4}$, weight decay $10^{-5}$, mixed precision, and lesion-containing patch sampling probability 0.7. No data augmentation was applied.

Each architecture was trained separately with Dice-BCE and Focal Tversky-Focal (FTF) losses. For lesion probabilities $p_i$, labels $g_i$, and $V$ voxels, Dice-BCE equally weighted soft Dice and binary cross-entropy:
\begin{align}
\mathcal{L}_{\mathrm{Dice-BCE}}
=&\ 0.5\left(1-\frac{2\sum_i p_i g_i+\epsilon}
{\sum_i p_i+\sum_i g_i+\epsilon}\right)\nonumber\\
&- \frac{0.5}{V}\sum_i\left[g_i\log p_i+(1-g_i)\log(1-p_i)\right],
\end{align}
with $\epsilon=10^{-6}$. FTF was
\begin{equation}
\mathcal{L}_{\mathrm{FTF}}=
0.5(1-\mathrm{TI})^{1/\gamma_T}+0.5\mathcal{L}_{\mathrm{Focal}},\quad
\mathrm{TI}=\frac{\mathrm{TP}+\epsilon}
{\mathrm{TP}+\alpha_T\mathrm{FP}+\beta_T\mathrm{FN}+\epsilon}.
\end{equation}
We used $\alpha_T=0.7$, $\beta_T=0.3$, $\gamma_T=1.33$, and $\epsilon=10^{-5}$; sigmoid focal loss used lesion weighting $\lambda=0.25$ and exponent $\gamma_F=2.0$. Logit-based implementations were used for numerical stability.

\subsection{Evaluation and Statistical Analysis}

Dice, sensitivity, and precision were calculated for all 85 FCD subjects. Overall Dice assigned zero to cases without lesion overlap; detected Dice was calculated among successfully localised cases (Dice $>0$). Miss rate was the proportion with Dice $=0$. Localisation was also summarised by lesion-volume quartile.

The six model-loss configurations were compared using the Friedman test, followed by two-sided paired Wilcoxon signed-rank tests with Benjamini-Hochberg correction across 15 comparisons and matched-pairs rank-biserial effect sizes. For CRIL-FTF, logistic regression examined successful localisation. Lesion volume was log-transformed to base 2, and a multivariable model adjusted for age and frontal involvement. Spearman correlations examined volume versus overall Dice, detected Dice, and within-subject improvement over 3D U-Net. Mann-Whitney and Kruskal-Wallis tests assessed exploratory clinical subgroups, with correction applied within analysis families.

\section{Results}

\subsection{Segmentation Performance}

Table~\ref{tab:main_results} summarises the six configurations. FTF improved performance for every architecture. CRIL-FTF achieved the highest overall Dice ($0.196\pm0.262$), sensitivity ($0.167\pm0.242$), precision ($0.397\pm0.424$), and detected Dice ($0.379\pm0.252$). It localised 44 of 85 lesions and reduced the miss rate to 48.2\%, compared with 57.6\% for 3D U-Net and 54.1\% for Self-Attention U-Net. Dice-BCE configurations had similar overall Dice ($0.083$-$0.096$) and missed approximately 71\% of lesions.

\begin{table}[t]
\centering
\scriptsize
\setlength{\tabcolsep}{2.2pt}
\caption{FCD segmentation performance (mean $\pm$ standard deviation; $n=85$). Dice $>0$ is detected-case Dice, with the number of localised cases in parentheses. Attn denotes Self-Attention U-Net.}
\label{tab:main_results}
\begin{tabular}{@{}llccccc@{}}
\toprule
Model & Loss & Dice & Sens. & Prec. & Dice $>0$ & Miss (\%)\\
\midrule
3D U-Net & Dice-BCE & $0.085\pm0.183$ & $0.074\pm0.178$ & $0.177\pm0.339$ & $0.302\pm0.230$ (24) & 71.8\\
3D U-Net & FTF & $0.136\pm0.224$ & $0.124\pm0.231$ & $0.301\pm0.390$ & $0.320\pm0.244$ (36) & 57.6\\
Attn & Dice-BCE & $0.083\pm0.157$ & $0.089\pm0.181$ & $0.116\pm0.232$ & $0.281\pm0.169$ (25) & 70.6\\
Attn & FTF & $0.135\pm0.214$ & $0.149\pm0.252$ & $0.190\pm0.294$ & $0.294\pm0.230$ (39) & 54.1\\
CRIL & Dice-BCE & $0.096\pm0.194$ & $0.078\pm0.161$ & $0.192\pm0.352$ & $0.341\pm0.222$ (24) & 71.8\\
CRIL & FTF & $\mathbf{0.196\pm0.262}$ & $\mathbf{0.167\pm0.242}$ & $\mathbf{0.397\pm0.424}$ & $\mathbf{0.379\pm0.252}$ (44) & \textbf{48.2}\\
\bottomrule
\end{tabular}
\end{table}

The Friedman test showed an overall difference among configurations ($\chi^2(5)=42.27$, $p<0.001$). FTF significantly improved Dice over Dice-BCE for 3D U-Net ($q=0.005$, $r_{\mathrm{rb}}=0.592$), CRIL-U-Net ($q<0.001$, $r_{\mathrm{rb}}=0.658$), and Self-Attention U-Net ($q=0.014$, $r_{\mathrm{rb}}=0.495$). Under FTF, CRIL-U-Net outperformed 3D U-Net ($q=0.047$, $r_{\mathrm{rb}}=0.357$) and Self-Attention U-Net ($q=0.016$, $r_{\mathrm{rb}}=0.431$); the two comparison architectures did not differ ($q=0.751$). Architectural differences under Dice-BCE were not significant (all $q\geq0.666$).

\subsection{Subject-Level and Qualitative Findings}

\begin{figure}[t]
\centering
\includegraphics[width=0.70\textwidth]{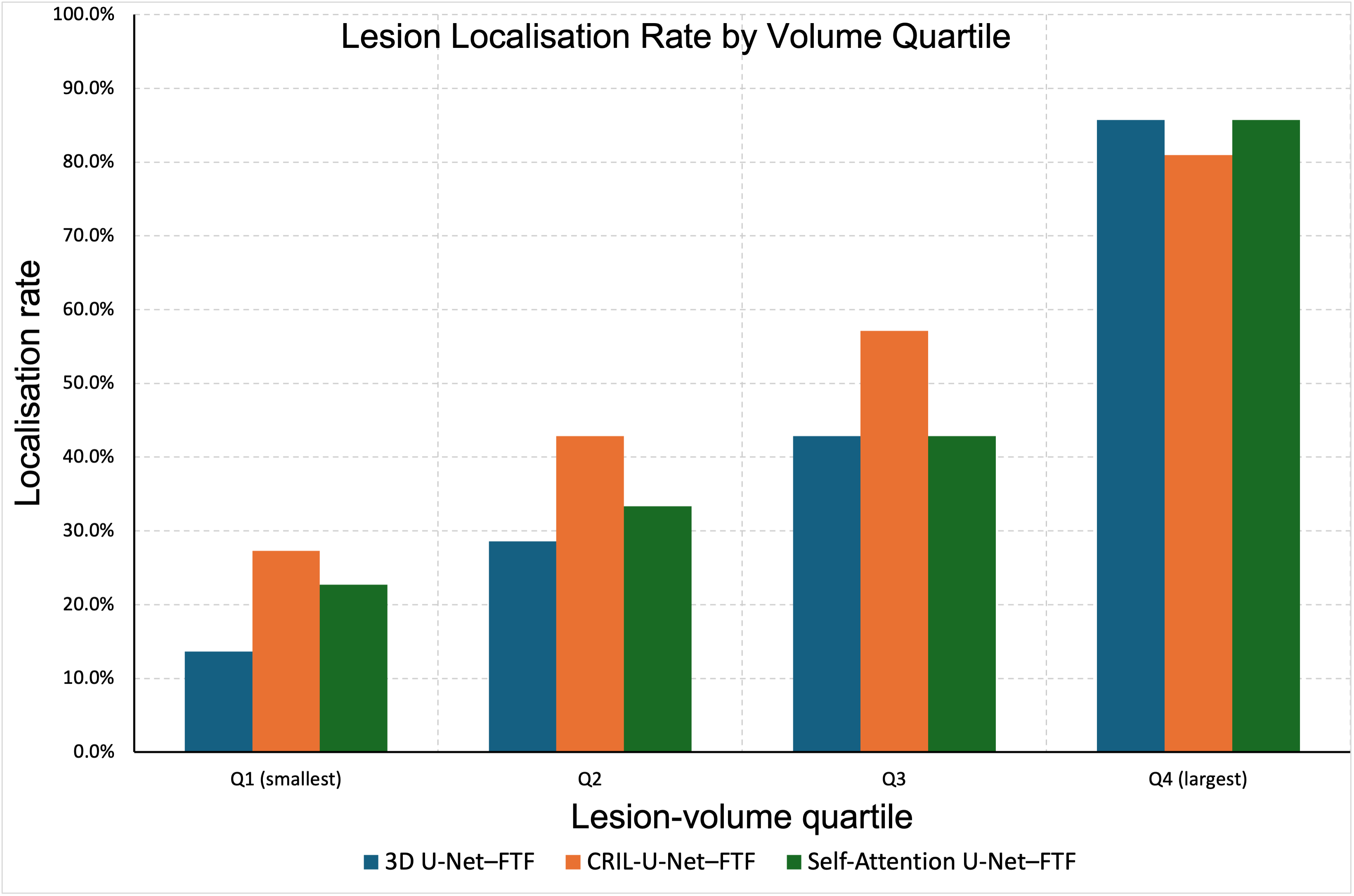}
\caption{Lesion localisation by ground-truth volume quartile for the three FTF models. CRIL-U-Net achieved numerically higher rates in Q1-Q3; all models performed similarly for the largest lesions.}
\label{fig:volume_localisation}
\end{figure}

Each doubling of lesion volume approximately doubled the odds of CRIL-FTF localisation (OR $=2.06$, 95\% CI $1.35$-$3.14$, $q=0.005$); the association remained after adjustment for age and frontal involvement (adjusted OR $=1.96$, 95\% CI $1.24$-$3.09$, $p=0.004$). Volume correlated with overall CRIL Dice ($\rho=0.342$, $q=0.008$), but not detected Dice ($\rho=-0.058$, $q=0.887$) or improvement over 3D U-Net ($\rho=-0.151$, $q=0.502$). As shown in Fig.~\ref{fig:volume_localisation}, CRIL localisation increased from 27.3\% in the smallest lesion-volume quartile to 81.0\% in the largest and was numerically higher than both comparators in the three smallest quartiles. Age, onset age, sex, hemisphere, and frontal involvement were not significant after correction; all five MRI-negative lesions were missed.

\begin{figure}[t]
\centering
\includegraphics[width=0.9\textwidth]{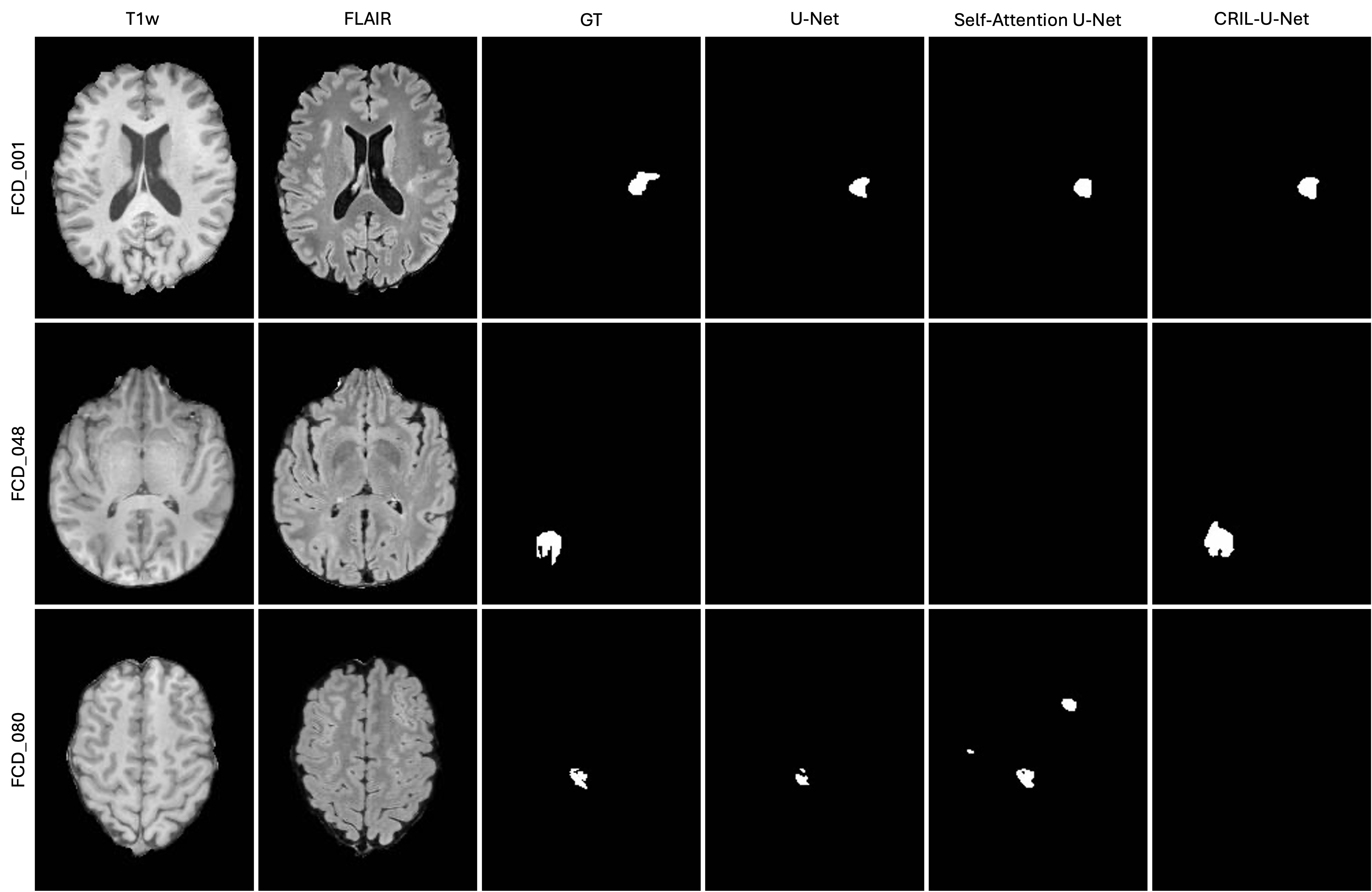}
\caption{Representative FTF results showing T1w, FLAIR, ground truth (GT), and predictions from 3D U-Net, Self-Attention U-Net, and CRIL-U-Net.}
\label{fig:qualitative_results}
\end{figure}

Figure~\ref{fig:qualitative_results} illustrates case-level variability. For FCD\_048, CRIL-U-Net achieved Dice $=0.642$, compared with 0.008 and 0.016 for 3D U-Net and Self-Attention U-Net. All models segmented FCD\_001 similarly, whereas CRIL missed FCD\_080, which the comparison models localised. Thus, the aggregate CRIL benefit did not extend to every subject.

\section{Discussion}

CRIL-U-Net with FTF achieved the best overall performance, increasing mean Dice from $0.136$ to $0.196$ and localising eight more lesions than 3D U-Net under identical experimental conditions. The improvement remained significant after false-discovery-rate correction. Because CRIL added only 4,320 parameters, the result is unlikely to be explained by a meaningful increase in model capacity. Instead, it supports explicitly learning local, voxel-wise cross-modal, and ratio-inspired relationships before segmentation.

The benefit depended strongly on the objective function. Under Dice-BCE, all architectures produced similar Dice scores and missed about 71\% of lesions. FTF significantly improved each architecture, and CRIL's advantage emerged only with this imbalance-aware objective. This interaction suggests that a useful representation is insufficient when optimisation remains dominated by the extensive background class. Conversely, self-attention did not significantly improve the common backbone under either loss and was inferior to CRIL under FTF. For subtle local abnormalities, directly modelling modality relationships may therefore be more effective than adding global context at the input.

Lesion volume chiefly affected whether a lesion was localised. Its association with overall Dice disappeared when analysis was restricted to localised cases, indicating that size influenced detection more than delineation quality after detection. CRIL showed its largest descriptive localisation advantage in the smaller-volume quartiles, although the within-subject improvement over 3D U-Net was not significantly correlated with volume. The missed MRI-negative cases and qualitative failures emphasise the remaining heterogeneity.
CRIL uses routinely acquired T1w and FLAIR MRI and requires no additional sequences. Nevertheless, a 48.2\% miss rate precludes independent clinical use. The method should be considered a compact research approach that may support future decision-assistance systems rather than a clinically deployable tool.

Limitations include the small single-dataset cohort, modest absolute performance, lack of external validation, and underpowered subgroup analyses. Component-wise ablation, direct comparison with fixed ratios, nnU-Net and modern transformer-based models, and augmentation were not evaluated. Future work should test the ratio branch separately, assess larger multicentre cohorts, and improve lesion-level localisation while preserving controlled comparisons.

\section{Conclusion}

CRIL-U-Net learns compact cross-modal representations from T1w and FLAIR before 3D U-Net segmentation. Under FTF loss it significantly outperformed direct concatenation and input self-attention, improving mean Dice and reducing lesion misses with only a 0.019\% parameter increase. The findings demonstrate the value of coupling explicit cross-modal representation learning with imbalance-aware optimisation, while the remaining miss rate motivates further validation and development.

\paragraph{Code Availability.}
The source code and experimental configurations are publicly available at
\url{https://github.com/soumenca/FCD_LatentRatio}.

\bibliographystyle{splncs04}
\bibliography{ref}

\end{document}